\documentclass[11pt]{article}
\pdfoutput=1
\usepackage[
  margin=0.83in,
  headheight=12pt,
  headsep=25pt,
  includefoot,
  footskip=30pt,
]{geometry}
\usepackage{tikz}
\usepackage{graphicx}
\usepackage{comment}
\usepackage{listings}
\usepackage{xcolor}
\usepackage{inconsolata}
\usepackage{makecell}
\usepackage{tabularx} 
\usepackage{adjustbox}
\usepackage{rotating}
\usepackage{pifont}

\usepackage{natbib}
\renewcommand\cite{\citep}  

\newcommand*\colourcross[1]{%
  \expandafter\newcommand\csname #1cross\endcsname{\textcolor{#1}{\ding{56}}}%
}
\colourcross{red}
\newcommand*\colourcheck[1]{%
  \expandafter\newcommand\csname #1check\endcsname{\textcolor{#1}{\ding{52}}}%
}
\colourcheck{green}

\definecolor{commentcolour}{rgb}{0.3,0.7,0.2}
\definecolor{backcolour}{rgb}{0.98,0.98,0.98}

\lstdefinelanguage{markdown}{
    comment=[l]{\#},
    morestring=[s]{```}{```},
    commentstyle=\color{commentcolour}\bfseries,
    stringstyle=\color{blue},
    basicstyle=\scriptsize\ttfamily,
    showstringspaces=false,
    breaklines=true,
    breakautoindent=false,
    breakindent=0pt,
    backgroundcolor=\color{backcolour},
}
\lstdefinestyle{mystyle}{
    morekeywords={self},
    basicstyle=\scriptsize\ttfamily,
    keywordstyle=\color{blue},
    commentstyle=\color{commentcolour}\bfseries,
    breaklines=true,
    breakautoindent=false,
    showstringspaces=false,
    backgroundcolor=\color{backcolour},
    stringstyle=\color{red},
}
\lstdefinelanguage{PythonPlus}[]{Python}{
  alsoother={@},
  morekeywords=[1]{,as,assert,nonlocal,with,yield,self,True,False,None} 
  morekeywords=[2]{,__init__,__add__,__mul__,__div__,__sub__,__call__,__getitem__,__setitem__,__eq__,__ne__,__nonzero__,__rmul__,__radd__,__repr__,__str__,__get__,__truediv__,__pow__,__name__,__future__,__all__,}, 
  morekeywords=[3]{,object,type,isinstance,copy,deepcopy,zip,enumerate,reversed,list,set,len,dict,tuple,range,xrange,append,execfile,real,imag,reduce,str,repr,}, 
  morekeywords=[4]{,Exception,NameError,IndexError,SyntaxError,TypeError,ValueError,OverflowError,ZeroDivisionError,}, 
  morekeywords=[5]{,ode,fsolve,sqrt,exp,sin,cos,arctan,arctan2,arccos,pi, array,norm,solve,dot,arange,isscalar,max,sum,flatten,shape,reshape,find,any,all,abs,plot,linspace,legend,quad,polyval,polyfit,hstack,concatenate,vstack,column_stack,empty,zeros,ones,rand,vander,grid,pcolor,eig,eigs,eigvals,svd,qr,tan,det,logspace,roll,min,mean,cumsum,cumprod,diff,vectorize,lstsq,cla,eye,xlabel,ylabel,squeeze,}, 
}

\usepackage{tikz}

\usepackage{amsmath} 
\usepackage{array} 
\usetikzlibrary{matrix, arrows}
\usepackage{amsmath,amssymb}
\usepackage{amsthm}
\usepackage{mathtools}
\usepackage{xspace}
\usepackage[noend]{algorithmic}
\usepackage[ruled,vlined]{algorithm2e}
\usepackage{url}
\usepackage{makeidx}
\usepackage{enumerate}
\usepackage{epstopdf}
\usepackage{booktabs}
\usepackage{color}
\usepackage[utf8]{inputenc}
\usepackage{thm-restate}
\usepackage{scalerel,stackengine}
\usepackage[shortlabels]{enumitem}
\usepackage{xr}
\usepackage{fancyvrb}
\usepackage{xcolor}
\usepackage{bold-extra}
\usepackage{arydshln}
\usepackage[small]{caption}
\usepackage{subcaption}
\usepackage[most]{tcolorbox}
\usepackage{fvextra}
\usepackage{float}
\usepackage{alltt}
\usepackage{soul}
\usepackage{MnSymbol,wasysym}
\usepackage{fancyvrb}
\usepackage{multirow}
\usepackage[bottom]{footmisc}
\usepackage[hidelinks]{hyperref}
\usepackage{diagbox}
\usepackage{mdframed}
\usepackage{todonotes}
\setuptodonotes{inline}
\usepackage{tikz}
\usetikzlibrary{shapes,calc,positioning}

\global

\tcbset{
  aibox/.style={
    width=\textwidth,
    top=0pt, bottom=0pt, left=5pt, right=5pt,
    colback=white,
    colframe=black,
    colbacktitle=black,
    enhanced,
    center,
    attach boxed title to top left={yshift=-0.1in,xshift=0.15in},
    boxed title style={boxrule=0pt,colframe=white,},
  }
}
\newtcolorbox{AIbox}[2][]{aibox,title=#2,#1}

\definecolor{aigold}{RGB}{244,210, 1} 
\definecolor{aigreen}{RGB}{210,244,211} 

\sethlcolor{aigreen}

\definecolor{aired}{RGB}{255,180,181}

\newtcbox{\mybox}[1][green]{on line,
arc=0pt,outer arc=0pt,colback=#1!10!white,colframe=#1!50!black,
boxsep=0pt,left=0pt,right=0pt,top=0pt,bottom=0pt,
boxrule=0pt,bottomrule=0pt,toprule=0pt}

\newcommand{\phil}{\texttt{Phi-4-Mini}\xspace}
\newcommand{\phir}{\texttt{Phi-4-Mini-Reasoning}\xspace}

\newcommand\extrafootertext[1]{%
    \bgroup
    \renewcommand\thefootnote{\fnsymbol{footnote}}%
    \renewcommand\thempfootnote{\fnsymbol{mpfootnote}}%
    \footnotetext[0]{#1}%
    \egroup
}
\begin{document}

\title{Phi-4-Mini-Reasoning: Exploring the Limits of Small Reasoning Language Models in Math}

\author{Haoran Xu$^\ddagger$~
Baolin Peng$^\ddagger$~
Hany Awadalla~
Dongdong Chen~
Yen-Chun Chen~
Mei Gao~\\
Young Jin Kim~
Yunsheng Li~
Liliang Ren~
Yelong Shen~
Shuohang Wang~
Weijian Xu~\\
Jianfeng Gao~
Weizhu Chen
\\\\
Microsoft
}
\date{}
\extrafootertext{$^\ddagger$Equal Contribution. Except for the first and last two authors, the remaining authors are listed in alphabetical order.}

\maketitle

\begin{abstract}
Chain-of-Thought (CoT) significantly enhances formal reasoning capabilities in Large Language Models (LLMs) by training them to explicitly generate intermediate reasoning steps. While LLMs readily benefit from such techniques, improving reasoning in Small Language Models (SLMs) remains challenging due to their limited model capacity. Recent work by Deepseek-R1 \citep{deepscaler2025} demonstrates that distillation from LLM-generated synthetic data can substantially improve the reasoning ability of SLM. However, the detailed modeling recipe is not disclosed. In this work, we present a systematic training recipe for SLMs that consists of four steps: (1) large-scale mid-training on diverse distilled long-CoT data, (2) supervised fine-tuning on high-quality long-CoT data, (3) Rollout DPO leveraging a carefully curated preference dataset, and (4) Reinforcement Learning (RL) with Verifiable Reward. We apply our method on \phil, a compact 3.8B-parameter model. The resulting \textbf{\phir} model exceeds, on math reasoning tasks, much larger reasoning models, e.g., outperforming DeepSeek-R1-Distill-Qwen-7B by 3.2 points and DeepSeek-R1-Distill-Llama-8B by 7.7 points on Math-500.
Our results validate that a carefully designed training recipe, with large-scale high-quality CoT data, is effective to unlock strong reasoning capabilities even in resource-constrained small models.
\end{abstract}

\begin{figure}[H]
    \centering
    \resizebox{0.76\linewidth}{!}{
    \includegraphics[width=7.5cm]{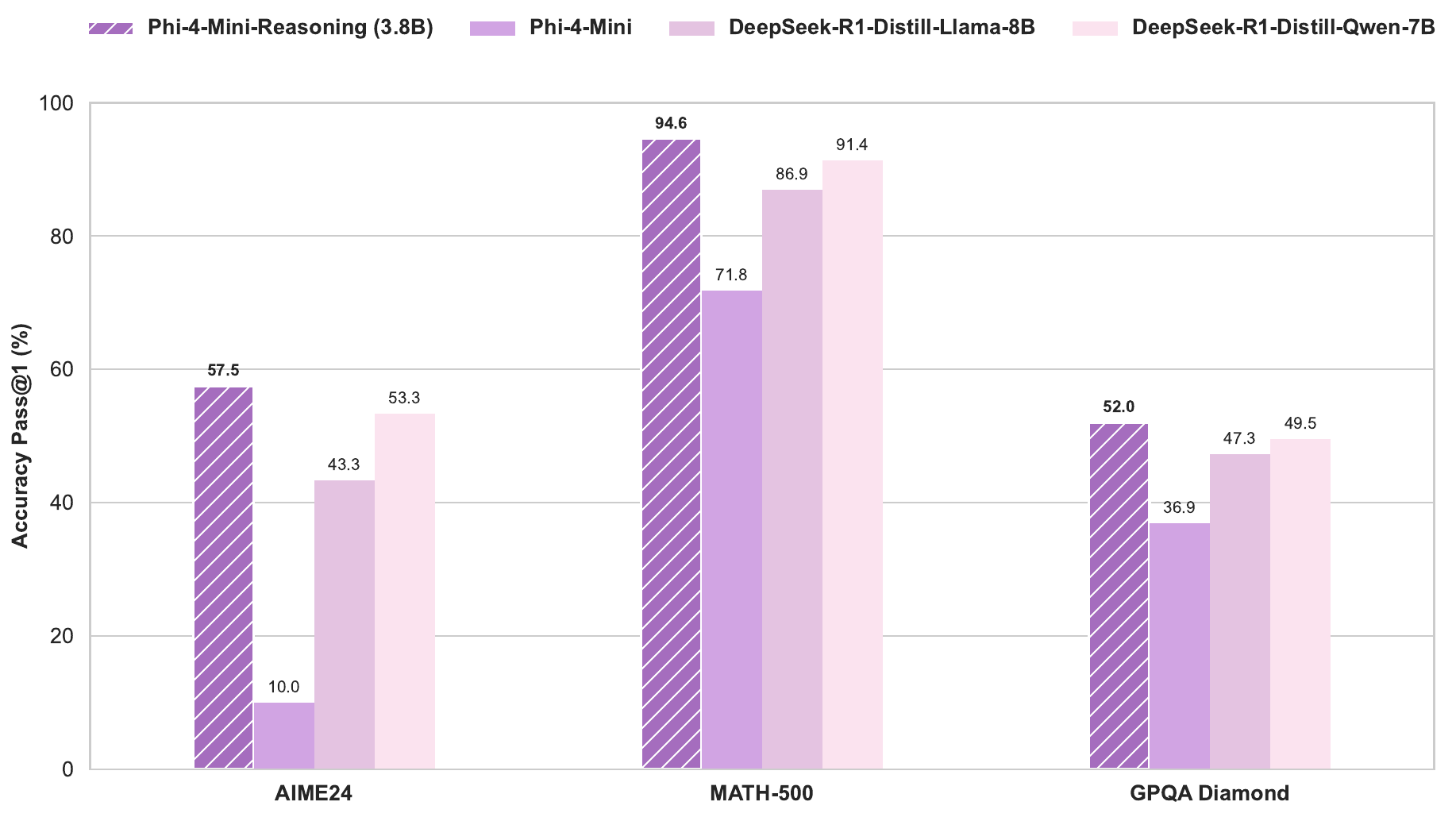}}
    \caption{Math benchmark performance of Phi-4-Mini-Reasoning.
    }
    \label{fig:overview}
\end{figure}
\section{Introduction}
Large Language Models (LLMs) have demonstrated remarkable capabilities across numerous natural language processing tasks, while their reasoning ability often deteriorates when confronting intricate, multi-step problems, where simply outputting an answer without intermediate steps leads to significant performance gaps \citep{wei2022chain}. The Chain-of-Thought (CoT) approach addresses this challenge by explicitly prompting models to generate a sequence of logical steps prior to arriving at a final answer, thereby significantly enhancing their reasoning capacities \cite{kojima2022large,wei2022chain}. Incorporating this reasoning process during inference has established the paradigm of test-time scaling, which further elevates performance in complex reasoning tasks \citep{snell2024scaling, welleck2024decoding, openai_o1}.

Enhancing reasoning abilities is inherently easier for larger LLMs due to their extensive capacity, whereas it remains challenging for Small Language Models (SLMs). Fortunately, Deepseek-R1 \citep{deepseekr1} indicates that non-logits-level distillation—effectively supervised fine-tuning (SFT) of SLMs using synthetic data generated by more capable models—can markedly improve SLM reasoning performance. For instance, such an approach could elevate MATH-500 \cite{math-500} accuracy of Llama-8B \citep{grattafiori2024llama} from 44.4\% to 89.1\% . Following this breakthrough, numerous efforts, including Bespoke-Stratos-7B \citep{bespoke_stratos} and OpenThinker-7B \citep{openthoughts}, have aimed to replicate these results. Despite this enthusiasm, debates persist regarding the primary focus of training. Deepscaler \citep{deepscaler2025} suggests scaling RL like GRPO \citep{grpo} for reasoning gains, while S1 and LIMO \citep{s1,limo} emphasize the quality and diversity of reasoning datasets, revealing that even datasets as small as fewer than 1K examples can enhance reasoning performance. 

Rather than focusing on isolated techniques that individually benefit training, we systematically explore a training paradigm specifically tailored for SLMs, where limited model capacity makes reasoning improvements particularly challenging. Our methodology consists of two stages of distillation, followed by rollout-based preference learning that also reuses wrong LLM-generated samples, and concludes with RL using a verifiable reward. Initially, we employ distillation as a mid-training mechanism to embed foundational reasoning capabilities. We then apply distillation again in a fine-tuning phase to further improve model generalization. During LLM rollout sampling for distillation, some incorrect outputs are typically discarded; however, we re-purpose these discarded samples to create a customized preference dataset, which is used for preference learning applied on top of the distilled model. Finally, we fine-tune the model using reinforcement learning with a verifiable reward signal based on final answer correctness. To ensure stable training, we introduce several targeted improvements, including prompt optimization, reward re-balancing via oversampling and filtering, and temperature annealing during exploration. 

We validate our proposed approach using Phi-4-Mini \citep{phi4-mini}, a compact 3.8-billion-parameter model, resulting in \textbf{\phir}, which outperforms other reasoning models nearly twice its size, such as DeepSeek-R1-Distill-Qwen-7B and DeepSeek-R1-Distill-Llama-8B.

\section{Background}
Small language models (SLMs) have demonstrated significant potential for strong reasoning capabilities. For example, Qwen-1.5B can achieve 83.9\% accuracy on the Math-500 \citep{math-500} benchmark simply by distilling 800K examples from DeepSeek-R1 \citep{deepseekr1}. Distillation has emerged as a powerful tool to enhance the reasoning abilities of SLMs; however, the optimal distillation strategy for small models remains underexplored. Recent studies provide complementary insights: \citet{deepscaler2025} suggests that gradually increasing generation length via reinforcement learning can further improve distilled models, while \citet{s1} and \citet{limo} emphasize that data diversity and quality, rather than quantity alone, are critical to success. Despite these advances, a comprehensive understanding of an effective distillation recipe for SLMs is still lacking. Moreover, naively applying isolated techniques can lead to degraded performance. For instance, directly distilling S1K \citep{s1} or LIMO \citep{limo} datasets onto Phi-4-Mini results in a significant drop in reasoning performance. This observation suggests that SLMs, due to their limited capacity, require substantially more carefully designed data and training strategies to develop robust reasoning capabilities compared to their larger counterparts. Detailed results illustrating this phenomenon are shown in Table \ref{tab:phi4_limo}.

  \begin{table}[h]
      \centering
      \resizebox{0.75\linewidth}{!}
      {
      \begin{tabular}{lccc}
      \toprule
      \textbf{Model} & \textbf{AIME 2024} & \textbf{MATH-500} & \textbf{GPQA Diamond} \\
      \midrule
      Phi-4-Mini & 10.0  & 71.8      & 36.9   \\
      Phi-4-Mini + LIMO & 6.7 & 57.8 &	24.8 \\
      Phi-4-Mini + S1K &3.0	&47.0	& 26.3 \\
        \midrule
      Phi-4-Mini-Reasoning (with our full recipe) & \textbf{57.5}  & \textbf{94.6} & \textbf{52.0} \\
      \bottomrule
      \end{tabular}
      }
      \caption{Pass@1 performance of Phi-4-Mini under different distillation settings. Naively using a small amount of high-quality data leads to significant performance degradation, highlighting the necessity of a comprehensive training recipe.}
      \label{tab:phi4_limo}
      \end{table}
      
Hence, our goal is to develop a comprehensive and efficient recipe for training SLMs. We first observe that non-reasoning SLMs require an initial mid-training stage to absorb a large volume of reasoning trajectories before any additional techniques are applied. However, several key questions remain: How much mid-training data is necessary? What subsequent techniques--such as careful distillation, preference learning, or reinforcement learning--should be employed next? In this work, we systematically address these questions and propose a complete recipe for building high-performing reasoning SLMs.

\section{Multi-Stage Continual Training for Reasoning}
Here, we systematically introduce our complete training recipe rather than exploring individual components. Taking a pre-trained SLM as a base, we improve its formal reasoning capability by first performing a multi-stage continual training using a curated CoT reasoning dataset and then running RL with verifiable rewards.
\subsection{Distillation as Mid-Training}
In the first stage, we frame distillation as mid-training. Specifically, we train the base model with next token prediction on an extensive corpus of synthetic chain-of-thought (CoT) data, which covers questions from diverse domains and varying levels of difficulty. The CoT-style answers for these questions are sampled by the Deepseek-R1 model \citep{deepseekr1}, after which we apply rejection sampling to retain only the correct answers. More details on our data generation methodology are presented in Section \ref{sec:data}. We pair each question with its corresponding correct CoT answer and train the base model using the standard causal language modeling objective. We train the model under a \textit{packing} mode, i.e., multiple short examples are packed in the same input sequence to increase training efficiency. The goal of this mid-training step is to equip the small base model with general CoT reasoning capabilities that are not explicitly learned during model mid-training. We find it effective to allow mid-training to iteratively use as much CoT training data as possible until model performance saturates on a validation dataset.

\subsection{Distillation as Supervised Fine-tuning}
After learning extensive and diverse reasoning chains, our next step involves selecting a compact, yet representative, subset from the mid-training dataset for subsequent fine-tuning.  Fine-tuning is performed in a \textit{non-packing} mode where we teach the model to decide where to stop generating. 
As it has been shown that higher-quality data can notably improve model performance and generalization capabilities and enable the model to better answer complex questions \citep{alma,lima,limo,s1}, we have constructed a combined dataset spanning diverse math domains, with difficulty levels exceeding the `college level'. More details about data categorization are described in Section \ref{sec:data}.

\subsection{Rollout Preference Learning}
In the previous two stages, the model is trained exclusively on accepted generations, filtering out rollouts containing incorrect answers. However, are the rejected rollouts entirely devoid of value? In this stage, we use rejected rollouts to enhance model performance. The quality of rejected data is important for preference learning, as pointed out by \citet{cpo},. Specifically, incorrect responses with minor nuances compared to their correct counterparts provide effective candidates for constructing informative preference pairs. To ensure data quality, we retained the questions that are categorized as `high-school' level math or above, determined by GPT-4o-mini \citep{gpt4}. The preference dataset is then constructed by designating correct answers as preferred rollouts and incorrect answers as dis-preferred rollouts for each question. Finally, we apply Direct Preference Optimization (DPO) \citep{dpo} to the model:

\begin{equation}
J_{\mathrm{DPO}}(\pi_\theta;\pi_{\text{ref}}) =  -\mathbb{E}_{(x,y_w,y_l) \sim \mathcal{D}} \Big[ \log \sigma \Big( \beta \log \frac{\pi_{\theta}(y_w | x)}{\pi_{\text{ref}}(y_w | x)}
 - \beta \log \frac{\pi_{\theta}(y_l | x)}{\pi_{\text{ref}} (y_l | x)} \Big) \Big],
\label{eq:dpo_loss}
\end{equation}
where $\pi_{\text{ref}}$ is the reference model, $y_w$ and $y_l$ are preferred and dis-preferred rollouts, respectively.

\subsection{RL with Verifiable Reward}
Although DPO improves the model's alignment and reasoning ability using curated preference pairs, DPO is limited as an offline learning method using a fixed dataset. To improve model's reasoning capability through online learning, we perform RL on the distilled and preference-trained model.
In what follows, we describe the RL algorithms we have experimented and the RL training recipe.

\paragraph{Proximal Policy Optimization (PPO)}
PPO~\citep{schulman2017proximal} has been successfully applied to fine-tuning LLMs via RLHF. The algorithm employs a clipped surrogate objective to limit each policy update so that it stays close to the previous policy. This clipping mechanism avoids overly large importance sampling ratios, which both stabilizes learning and enhances sample efficiency. PPO seeks to maximize
\begin{equation}
J_{\mathrm{PPO}}(\theta)
= \mathbb{E}_{q\sim\mathcal{D},\,o_{\le t}\sim\pi_{\theta_{\mathrm{old}}}(\cdot\mid q)}
\!\biggl[\,
  \min\Bigl(
    r_t(\theta)\,\widehat{A}_t,\;
    \mathrm{clip}\bigl(r_t(\theta),\,1-\epsilon,\,1+\epsilon\bigr)\,\widehat{A}_t
  \Bigr)
\biggr],
\label{eq:ppo}
\end{equation}
where
\[
r_t(\theta)
= \frac{\pi_\theta(o_t\mid q,o_{<t})}
       {\pi_{\theta_{\mathrm{old}}}(o_t\mid q,o_{<t})},
\]
and $q$ is sampled from the data distribution $\mathcal{D}$, $\epsilon$ controls the clipping range, and $\widehat{A}_t$ denotes the advantage estimate at time step $t$. To compute $\widehat{A}_t$, PPO uses the Generalized Advantage Estimator (GAE)~\cite{schulman2015high}.  Given a value function \(V\) and a reward function \(R\), the estimator is
\begin{equation}
\widehat{A}_t^{\mathrm{GAE}(\gamma,\lambda)}
= \sum_{l=0}^{\infty} (\gamma\,\lambda)^l\,\delta_{t+l},
\label{eq:gae}
\end{equation}
with the temporal‐difference term
\begin{equation}
\delta_l
= R_l + \gamma\,V(s_{l+1}) - V(s_l),
\quad
0 \le \gamma,\lambda \le 1.
\label{eq:delta}
\end{equation}

\paragraph{Group‐based Relative Policy Optimization (GRPO)}

GRPO~\citep{grpo} estimates its baseline by comparing rewards within a batch of \(G\) model responses, reducing the critic's cost and improving model training stability.  Concretely, for each question \(q\), it samples a set of candidate responses \(G\) \(\{o_i\}_{i=1}^G\) under the old policy \(\pi_{\theta_{\rm old}}\), then computes their rewards \(\{R_i\}_{i=1}^G\).  The normalized advantage is computed as
\begin{equation}
A_i \;=\;\frac{R_i - \mathrm{mean}(R_1,\dots,R_G)}
                     {\mathrm{std}(R_1,\dots,R_G)},
\label{eq:grpo-advantage}
\end{equation}

\noindent GRPO then maximizes a clipped‐surrogate objective, averaged over the group, with an additional KL‐penalty toward a reference policy \(\pi_{\rm ref}\):
\begin{equation}
J_{\mathrm{GRPO}}(\theta)
= \mathbb{E}_{%
  q\sim \mathcal{D},\;\{o_i\}\sim\pi_{\theta_{\rm old}}(\cdot\mid q)}
\!\Biggl[
  \frac{1}{G}\sum_{i=1}^G
    \min\!\Bigl(
      r_i(\theta)\,A_i,\;
      \mathrm{clip}\bigl(r_i(\theta),\,1-\epsilon,\,1+\epsilon\bigr)\,A_i
    \Bigr)
  \;-\;\beta\,D_{KL}\!\bigl(\pi_\theta\Vert\pi_{\rm ref}\bigr)
\Biggr],
\label{eq:grpo-objective}
\end{equation}
where \(\epsilon\) is the clipping parameter and \(\beta\) weights the KL‐penalty.

\paragraph{Verifiable Reward}

Reinforcement learning with verifiable reward (RLVR) has shown to be very effective in training models for various reasoning tasks~\citep{deepseekr1,ye2025emergence}. Following prior work~\citep{deepseekr1}, the reward for a verifiable task is defined as a function of the accuracy of the model's final answer. Concretely, 

\begin{equation}\label{eq:rule_based_reward_paraphrase}
R(\hat y, y) =
\begin{cases}
+1, & \text{if }\texttt{verify}(\hat y, y),\\
-1, & \text{otherwise.}
\end{cases}
\end{equation}

\noindent where \(y\) denotes the ground-truth answer and \(\hat y\) the response of the model.

\paragraph{Our RL Recipe}

In our pilot study of applying GRPO to train our base model, we have observed three issues that affect the stability and effectiveness of model training.

\begin{enumerate}
  \item \textbf{High Variance in Response Lengths}  
  Although the base model, after mid-training, is already able to generate reasonable CoT responses, we have observed substantial variability in response lengths within the same GRPO sampling group. For the same prompt, positively rewarded responses ranged from approximately 12k to 20k tokens. Directly optimizing the model for the standard GRPO objective on such length-heterogeneous responses induces instability. \citet{zhang2025srpo} reports a similar phenomenon when training a model on both mathematical and coding tasks.

  \item \textbf{Vanishing Gradients under Uniform Rewards}  
  GRPO’s reliance on advantage estimates makes it susceptible to the vanishing gradient problem of all sampled responses in a group receiving identical rewards, yielding zero variance in the returns. The DAPO framework~\citep{yu2503dapo} addresses this problem by oversampling and filtering out prompts whose response accuracies are exactly 0 or 1, thereby preserving non-zero advantage signals. However, we find in our experiments that we need to address the following two problems when applying DAPO to our model:
  \begin{enumerate}[(i)]
    \item The model is sensitive to intra-group length discrepancies: responses with intermediate accuracies (e.g., 0.1 or 0.9) still provoke unstable gradient magnitudes due to response-length variance.
    \item For difficult math tasks, attaining even a single positively rewarded sample (by prompting the model) requires expanding the GRPO batch size to 128. This imbalance between positive and negative training signals impedes RL convergence.
  \end{enumerate}
  We hypothesize that these issues become more prominent for small language models, where the RL stability is more likely to be fragile, compared to LLMs.

  \item \textbf{Exploration–Exploitation Tradeoff}  
  Effective exploration is essential for discovering high-reward policies in RL. While a sampling temperature of 1.0 or higher is employed to encourage explorations, a lower temperature (e.g. 0.6) is typically used to constrain output variance on math and coding tasks. In our experiments, we have observed a substantial performance gap resulting from this divergence between the exploration used during training and the exploitation settings applied at evaluation.
\end{enumerate}

To address the aforementioned challenges, we introduce a set of methods to improve the stability and effectiveness of RL training:

\begin{enumerate}
    \item \textbf{Prompt Optimization} 
    We perform multiple rounds of sampling using multiple candidate prompts intended for RL training using the distilled model. Then only those prompts whose generated responses exhibit relatively uniform token lengths are retained. This method mitigates the instability induced by high intra-group response length variance during GRPO optimization.
    
    \item \textbf{Reward Rebalancing through Oversampling and Filtering}
    Inspired by DAPO~\citep{yu2503dapo}, for difficult prompts, we first conduct oversampling to ensure sufficient diversity in the response group. We then re-balance the group by retaining all positive-reward responses and randomly sampling an equal number of negative-reward responses. To further reduce the length variance and avoid instability from overly easy prompts, we filter out prompts whose group-level accuracy exceeds a certain threshold (e.g., 50\%).
    
    \item \textbf{Temperature Annealing}
   To seek the best tradeoff between exploration and exploitation during the course of model training, we introduce temperature annealing. We initialize the sampling temperature as 1.0 and linearly decay it over the first 50\% of training steps down to 0.6. For the remaining training steps, the temperature is fixed as 0.6. This strategy encourages broader exploration in the early stage of RL while gradually transitioning toward the exploitation in the well-known state-action subspace. 
\end{enumerate}

\begin{table}[h]

\begin{center}

\resizebox{0.6\linewidth}{!}{
\begin{tabular}{lcc}
\toprule
Data Resource       & Size & Reasoning \\ \midrule
AquaRAT \citep{aquarat}             & 98K  & \redcross        \\
Ape210K \citep{zhao2020ape210k}             & 210K & \redcross        \\
MetaMathQA \citep{yu2023metamath}          & 395K & \redcross        \\
MathInstruct \citep{yue2023mammoth}       & 262K & \redcross        \\
TAL-SCQ5K \citep{TALSCQ5K}          & 5K   & \redcross        \\
OpenR1-Math \citep{openr1}         & 220K & \greencheck       \\
Bespoke-Stratos-17k \citep{bespoke_stratos} & 17K  & \greencheck       \\
OpenThoughts-114K \citep{openthoughts}  & 114K & \greencheck       \\
\bottomrule
\end{tabular}
}

\end{center}

\caption{Overview of the data resources used for constructing the reasoning dataset. For non-reasoning data, we only use the questions and sample answers from Deepseek R1.}

\label{tab:data}

\end{table}
\section{Synthetic CoT Data Generation}
\label{sec:data}
To support distillation and rollout-based preference learning, we construct a large-scale reasoning dataset composed of LLM-generated synthesized reasoning trajectories. Specifically, we aggregate multiple public datasets—such as Bespoke \citep{bespoke_stratos}, Openthoughts \citep{openthoughts}, and OpenR1-Math \citep{openr1}—along with several in-house seed datasets. For datasets that already include reasoning trajectories, we directly use the provided annotations. For datasets lacking such trajectories, we retain only the math questions and generate new chain-of-thought answers using DeepSeek-R1 (671B). For each question, we sample approximately eight rollouts. An overview of the data sources is provided in Table~\ref{tab:data}. In total, we collect around 10 million rollouts across 1.6 million samples, including contributions from public datasets. For math questions that are verifiable, we first apply a math-verification tool to assess the correctness of the answers. However, as automatic verification can sometimes fail to validate complex solutions—leading to false negatives—we additionally employ GPT-4o-mini to re-verify rollouts initially flagged as incorrect. To maintain dataset balance, we annotate each data sample with attributes including the domain category, the difficulty level, and the presence of repetitive patterns. Domain categories cover a wide range of areas such as algebra, geometry, theory, probability, and calculus. Difficulty levels are categorized as elementary school, middle school, high school, college, and graduate level. The mid-training phase leverages the full dataset, while subsequent training steps operate on selected subsets.

\section{Experiment}
\subsection{Evaluation}
We evaluate our model on three mathematical reasoning tasks: AIME24 \citep{aime}, Math-500 \citep{math-500}, and GPQA Diamond \citep{gpqa}. For evaluation, the generation parameters are set with a temperature of 0.6, $\text{top}_\text{p}$ of 0.95, and a maximum sequence length of 32K. For each task, we conduct 3 runs and report the average performance across these trials.

\subsection{Baselines}
We compare our \phir model with o1-mini and several leading open-source, small-scale reasoning models, including DeepSeek-R1-Distill-Llama-8B \cite{deepseekr1}, Bespoke-Stratos-7B \cite{bespoke_stratos}, and OpenThinker-7B \cite{openthoughts}.

\subsection{Training Settings} 
For the first two distillation stages, we use a batch size of 128, a learning rate of 1e-5, a total of 5 training epochs, and a warmup ratio of 0.1. During the first stage, the sequence length is set to 16K \textit{with} packing strategy, whereas in the second stage the sequence length is extended to 20K \textit{without} packing. For the Rollout DPO phase, we use a learning rate of 5e-7 for a single training epoch, with a sequence length of 16K. During the RL stage, a learning rate of 5e-7 and a sequence length of 25k are used to encourage model exploration.

\subsection{Results}
The overall results are presented in Table~\ref{tab:overall_results}. \phir, despite having only 3.8 billion parameters, outperforms all open-source baseline models, including those nearly twice its size. In addition, we provide an ablation study to demonstrate the contribution of each training stage to the performance of \phir.
\begin{table}[h]

\begin{center}

\resizebox{0.7\linewidth}{!}{
\begin{tabular}{l||ccc}
\toprule

Model                        & AIME          & MATH-500      & GPQA Diamond   \\\midrule
o1-mini*                      & 63.6          & 90.0          & 60.0           \\
DeepSeek-R1-Distill-Qwen-7B  & 53.3          & 91.4          & 49.5          \\
DeepSeek-R1-Distill-Llama-8B & 43.3          & 86.9          & 47.3          \\
Bespoke-Stratos-7B*           & 20.0          & 82.0          & 37.8           \\
OpenThinker-7B*               & 31.3          & 83.0          & 42.4           \\ 
Llama-3.2-3B-Instruct        & 6.7             & 44.4             & 25.3              \\
\midrule
\phil                           & 10.0  & 71.8      & 36.9         \\
\quad + Distill Mid-training        & 30.0  & 82.9      & 42.6            \\
\quad + Distill Fine-tuning         & 43.3  & 89.3      & 48.3            \\
\quad + Roll-Out DPO   & 50.0 & 93.6 & 49.0 \\
\quad + RL (\phir)  & \textbf{57.5} & \textbf{94.6} & \textbf{52.0} \\
\bottomrule
\end{tabular}
}

\end{center}

\caption{Pass@1 CoT Reasoning results of \phir compared with larger 7B reasoning models and OpenAI models. An asterisk (*) indicates results taken directly from the published reports, while the remaining results were reproduced in our work.}

\label{tab:overall_results}

\end{table}

\subsection{Ablations}
In this section, we conduct ablation studies to understand the impact of our distillation training on the model’s reasoning capability and compare the training stability of our RL recipe with DAPO.

\begin{figure}[ht]
  \centering
  \begin{subfigure}[t]{0.5\textwidth}
    \centering
    \includegraphics[width=\linewidth,height=6cm,keepaspectratio]{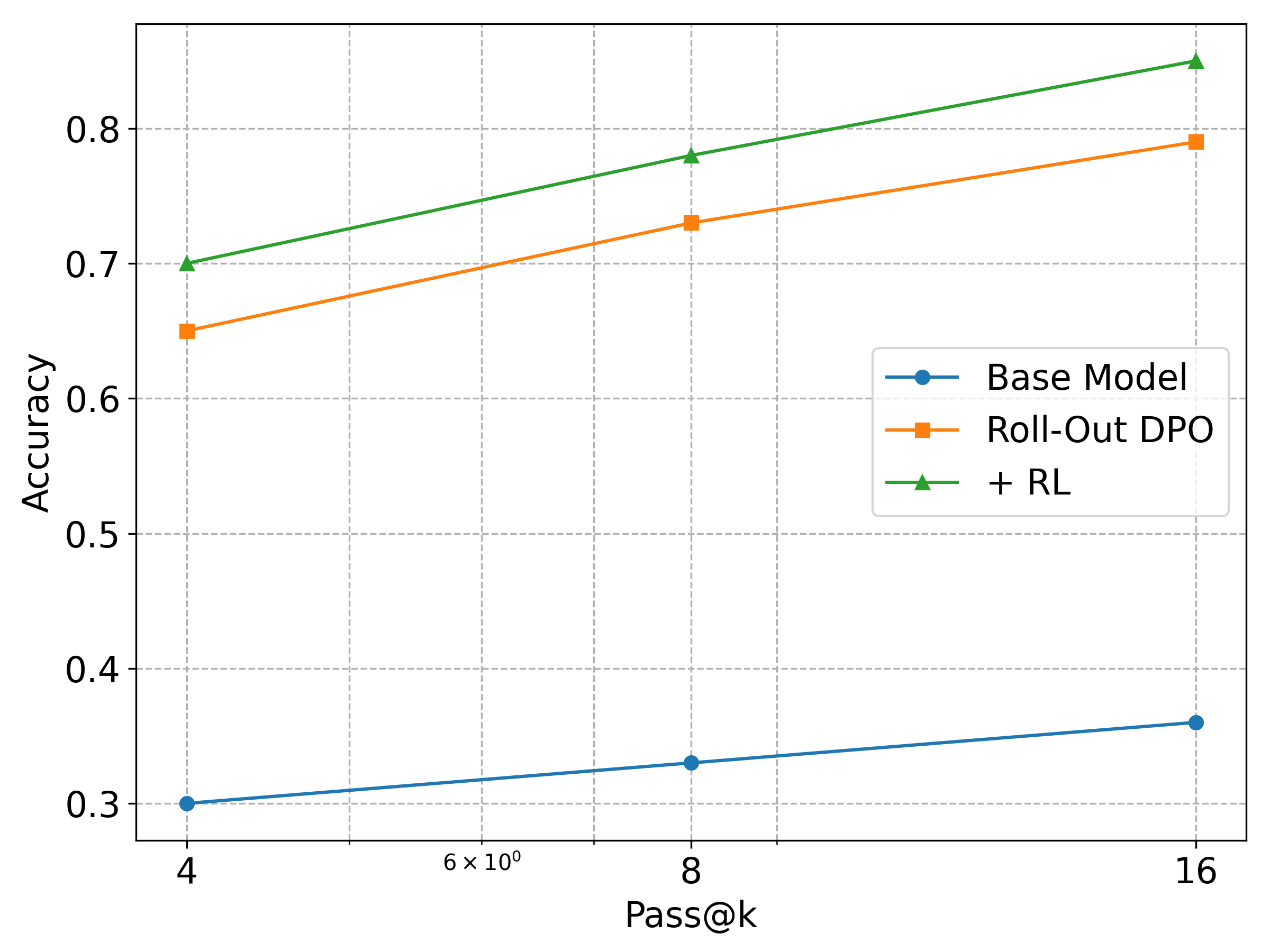}
    \caption{}
    \label{fig:passk}
  \end{subfigure}%
  \hfill
  \begin{subfigure}[t]{0.5\textwidth}
    \centering
    \includegraphics[width=\linewidth,height=6cm,keepaspectratio]{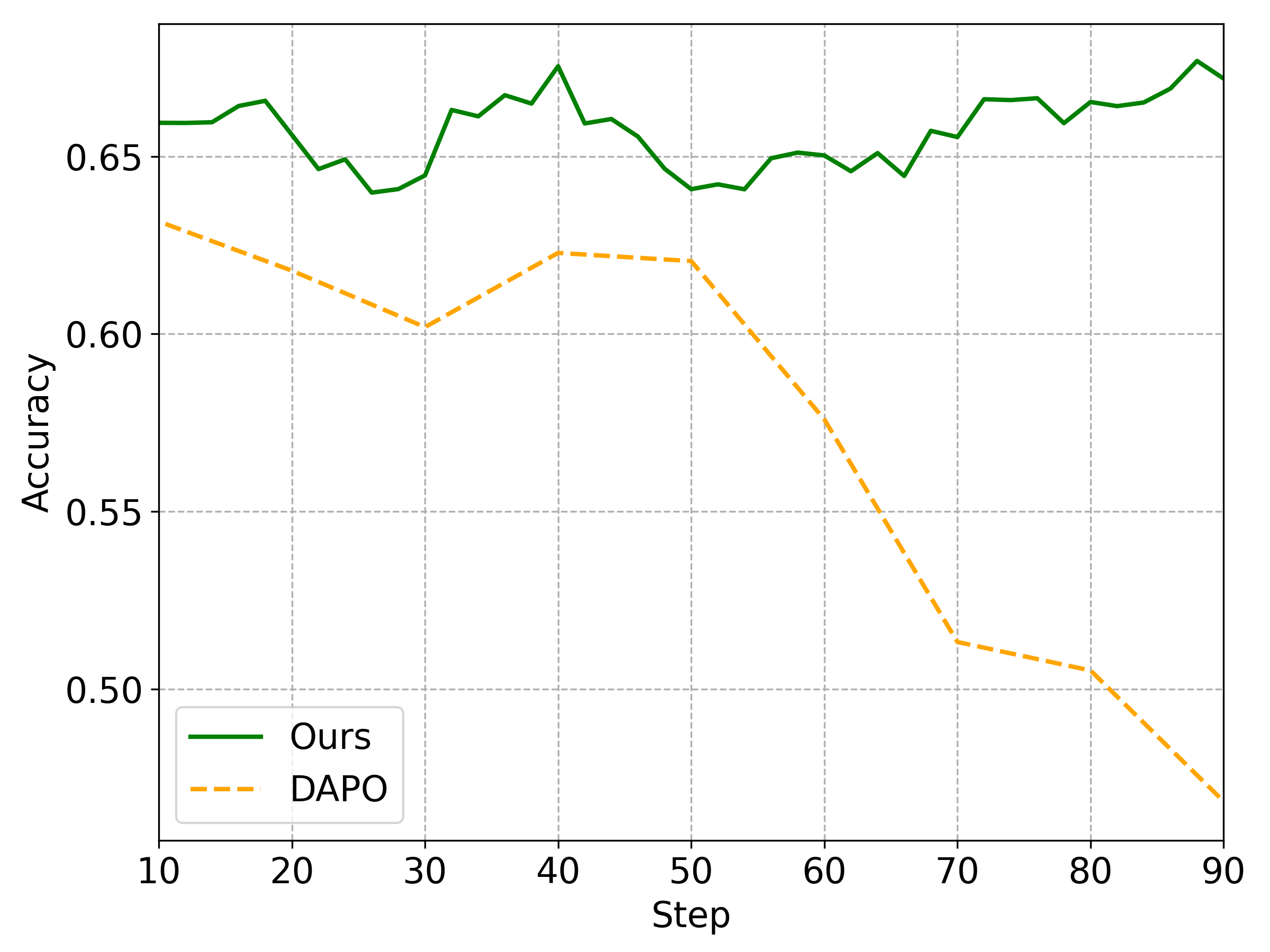}
    \caption{}
    \label{fig:rl_curve}
  \end{subfigure}

  \caption{%
    (a) Pass@k curves on AIME 2024 for the base model, the Roll-Out DPO model, and the model with additional RL training. Rollout DPO significantly improves Pass@k, extending the model’s reasoning capabilities. Further RL training yields additional gains.  
    (b) Comparison between DAPO and our RL training method, evaluated by cons@16 accuracy on AIME 2024. Our RL training approach demonstrates better stability.
  }
  \label{fig:sidebyside}
\end{figure}

To measure the reasoning boundary of an LLM, we use the pass@$k$ metric. For each problem, we sample $k$ outputs from the model. The pass@$k$ value for a question is 1 if at least one of the $k$ samples passes verification; otherwise, it is 0. The average pass@$k$ over the dataset reflects the proportion of problems that the model can solve within $k$ attempts. As shown in Figure~\ref{fig:passk}, our distillation pipeline serves as an effective approach for injecting reasoning-related knowledge into the model. After the distillation phase, pass@$k$ scores are substantially improved, indicating that distillation successfully extends the reasoning capability boundary of the base LLM. This lays a strong foundation for subsequent RL training. Building on this, RL fine-tuning further improves performance, providing an additional boost of approximately 7 points on average and further refining the model’s abilities.

We also compare our RL training method against DAPO. As shown in Figure~\ref{fig:rl_curve}, DAPO does not perform well in our setting: the consensus@16 metric on the AIME dataset consistently degrades as training progresses. In contrast, our RL training technique exhibits greater stability and consistently yields meaningful improvements over the base model.

\subsection{Safety Statement}
Phi-4-Mini-Reasoning was developed in accordance with Microsoft's responsible AI principles. Potential safety risks in the model’s responses were assessed using the Azure AI Foundry’s Risk and Safety Evaluation framework, focusing on harmful content, direct jailbreak, and model groundedness. The Phi-4-Mini-Reasoning Model Card contains additional information about our approach to safety and responsible AI considerations that developers should be aware of when using this model.

\section{Conclusion}
We present a multi-stage training paradigm to enhance reasoning capabilities in small language models (SLMs), combining large-scale distillation, rollout preference learning, and reinforcement learning with verifiable rewards. Applied to Phi-4-Mini, our approach produces \phir, a compact 3.8-billion-parameter model that outperforms open-source reasoning models nearly twice its size. We demonstrate that a carefully coordinated sequence of training stages is essential for unlocking robust reasoning in SLMs. Our results show that small models, when trained with deliberate data selection and training strategies, can match or even exceed the capabilities of much larger models. We believe that this work provides a blueprint for developing efficient, high-performing models under resource constraints.
\section*{Acknowledgments}
We extend our sincere gratitude to Amit Garg, Daniel Perez-Becker, Nguyen Bach, Tetyana Sych and the entire GenAI team for their invaluable contributions to this work. Their support in model review, deployment, and productization was instrumental in bringing this project to completion. We also gratefully acknowledge the Turing team for their ongoing technical collaboration and insightful discussions.

\clearpage
\bibliography{mainbib}

\end{document}